\documentclass[sn-mathphys,Numbered]{sn-jnl}

\usepackage{graphicx}%
\usepackage{multirow}%
\usepackage{amsmath,amssymb,amsfonts}%
\usepackage{amsthm}%
\usepackage{mathrsfs}%
\usepackage[title]{appendix}%
\usepackage{xcolor}%
\usepackage{textcomp}%
\usepackage{manyfoot}%
\usepackage{booktabs}%
\usepackage{algorithm}%
\usepackage{algorithmicx}%
\usepackage{algpseudocode}%
\usepackage{listings}%

\theoremstyle{thmstyleone}%
\theoremstyle{thmstyletwo}%

\theoremstyle{thmstylethree}%

\begin{document}

\title[Article Title]{A Comprehensive Study of Object Tracking in Low-Light Environments}


\author[1]{\fnm{Anqi} \sur{Yi}}\email{lm20690@bristol.ac.uk}

\author*[1]{\fnm{Nantheera} \sur{Anantrasirichai}}\email{n.anantrasirichai@bristol.ac.uk}

\affil*[1]{\orgdiv{Visual Information Laboratory}, \orgname{University of Bristol}, \orgaddress{\city{Bristol}, \postcode{BS8 1UB}, \country{UK}}}


\abstract{Accurate object tracking in low-light environments is crucial, particularly in surveillance and ethology applications. However, achieving this is significantly challenging due to the poor quality of captured sequences. Factors such as noise, color imbalance, and low contrast contribute to these challenges. This paper presents a comprehensive study examining the impact of these distortions on automatic object trackers. Additionally, we propose a solution to enhance tracking performance by integrating denoising and low-light enhancement methods into the transformer-based object tracking system. Experimental results show that the proposed tracker, trained with low-light synthetic datasets, outperforms both the vanilla MixFormer and Siam R-CNN.}

\keywords{tracking, low-light, enhancement, denoising}



\maketitle

\section{Introduction}\label{sec1}

The task of visual-based object tracking has been a core research area in computer vision for decades, focusing on determining the state of a designated target within video sequences, starting from its initial state. Its applications include surveillance, security, robotics, automotive, transportation, ethology, etc. However, tracking objects in low-light environments presents significant challenges due to the poor sequence quality captured. The presence of noise, motion blur, color imbalance, and low contrast in these sequences makes it difficult for traditional  algorithms to accurately track objects. This paper explores methods aimed at enhancing the performance of visual object tracking in low-light conditions and analyzing how various factors, such as noise, color imbalance, and low contrast, impact tracking effectiveness.

Similar to other computer vision tasks, deep learning has emerged as an effective tool for object tracking. Early learning-based methods adapted object recognition techniques to individual frames within a video \cite{Anantrasirichai:AI:2022}. Subsequently, recurrent neural networks (RNNs) were integrated to track detected objects over time. In 2017, the Transformer was introduced \cite{NIPS2017_3f5ee243}, proposing a novel architecture for natural language processing tasks that solely relies on attention mechanisms, eliminating the need for recurrent or convolutional neural networks (CNNs). The key innovation of the Transformer architecture is its ability to process input sequences in parallel, rather than sequentially as in traditional RNN models. This property enables more efficient training and faster convergence. Additionally, the Transformer model effectively handles long-range dependencies in input sequences, addressing a common issue faced by RNNs, which has made it an attractive option for object tracking.

This paper employs a state-of-the-art approach for transformer-based object tracking, MixFormer \cite{cui2022mixformer}, known for its superiority over several CNNs, RNNs, and earlier Transformer models.  MixFormer simplifies the traditional multi-stage pipeline and integrates feature extraction and target information integration within a unified transformer-based framework. In contrast to existing trackers, such as Siam R-CNN \cite{siamrcnn}, which relies on CNNs pretrained for generic object recognition, MixFormer leverages the flexibility and global modeling capacity of attention operations to capture target-specific features and promote wide range communication between the target and search area. By introducing a Mixed Attention Module (MAM) with hybrid interaction schemes, MixFormer enables simultaneous feature extraction and target integration, which results in a more compact and neat tracking pipeline. This approach overcomes the limitations of traditional trackers that use separate components for feature extraction, integration, and target-aware localization.

In low-light conditions, using the tracker faces several limitations:
\textit{i) Lack of specialized modules for low-light tracking:} The Mixformer is specifically designed to capture target features in daylight conditions. However, these features may become indistinct or distorted in low-light conditions due to insufficient lighting and noise, thereby limiting its performance.
\textit{ii) Limited training data in low-light conditions:} The model relies on a substantial amount of labelled data to effectively conduct visual object tracking. The inadequacy of training data specific to low-light conditions contributes to diminished performance in such scenarios. To address these issues, we propose integrating denoiser and enhancement module to the framework and using synthetic low-light datasets to train the model. 

This paper presents a comprehensive study on object tracking in low-light environments. We investigate the distinct types of distortions present in low-light content and their individual impacts on tracking performance. Subsequently, we enhance the trackers by employing preprocessing techniques involving denoising and brightness enhancement. Finally, we discuss the limitations of the current approach.

\section{Related work}\label{relatedwork}

\subsection{Object tracking}

The early work in learning-based object detection focused on fully convolutional networks (FCNs), which have demonstrated effectiveness in capturing both local and global contextual information during the tracking process \cite{wang2015visual}. Subsequently, more sophisticated methods gained popularity, such as the fully-convolutional Siamese network \cite{SiamFC}. Further enhancements include the Siamese Region Proposal Network (SiamRPN) \cite{SiamRPN}, which integrates the Siamese network with a region proposal mechanism for high-performance tracking. Additionally, the Discriminative Model Prediction (DiMP) \cite{bhat2019learning} was proposed to address object deformations and occlusions during tracking.

The transformer architecture and attention mechanisms have recently emerged as powerful techniques in various computer vision tasks, including object tracking. The transformer architecture has successfully replaced traditional convolutional layers with self-attention mechanisms. A well-known example of a transformer-based object tracking model is DETR (DEtection TRansformer)~\cite{carion2020end}, which captures both local and global context information, enabling it to handle complex scenarios effectively. Although DETR is primarily focused on object detection, it can be adapted for object tracking tasks by gathering information from multiple frames.  Another example is TrackFormer~\cite{meinhardt2022trackformer} for multi-object tracking. This single unified transformer architecture performs both detection and tracking in an end-to-end manner. The model demonstrates exceptional performance in multi-object tracking benchmarks. Similarly, the MOTR model \cite{zeng2022motr} employs a transformer-based architecture with temporal aggregation network for multiple object tracking. 

In addition to transformer-based models, attention mechanisms have been integrated into other object tracking models to enhance their performance. This includes the Distractor-aware Siamese Networks (DaSiamRPN)~\cite{zhu2018distractor}, where a distractor-aware training strategy  incorporates attention mechanisms to search objects effectively. This strategy improves the tracker's robustness against distractors. The Attentional Correlation Filter Network (ACFN), proposed in \cite{choi2017attentional}, incorporates an attention mechanism into a correlation filter-based tracker to adaptively weigh different spatial regions based on their importance during tracking.

\subsection{Low-light enhancement}

The advancements in deep learning have significantly progressed image enhancement, yet learning-based video enhancement remains relatively new. Some methods show promise in extending low-light image enhancement techniques to videos. These strategies involve estimating noise maps to guide attention mechanisms, implementing self-calibrated illumination frameworks, and utilizing adaptive total variation regularization (e.g., \cite{Xu:SNR-Aware:2022}, \cite{Ma:toward:2022}). Additionally, there's growing interest in techniques that integrate the Retinex theory model with learnable physical priors for reflectance and illumination (e.g., \cite{Wu:URetinex:2022}).

For video processing, various methods utilize alignment modules (e.g., \cite{zhou2021rta}) to synchronize feature maps of neighboring frames with the current frame, aiding in motion handling. Despite their purpose, these alignment modules occasionally fall short in compensating for motion adequately, leading to artifacts in feature combinations. Some approaches leverage Siamese Networks with shared weights to reduce noise in videos \cite{triantafyllidou2020low}. To cope with limited paired datasets, certain methods resort to unpaired training strategies, such as employing CycleGAN \cite{anantrasirichai:Contextual:2021}.

\section{Methods for object tracking in low-light environments}\label{sec2}

\begin{figure}[t]
\includegraphics[width=0.8\textwidth]{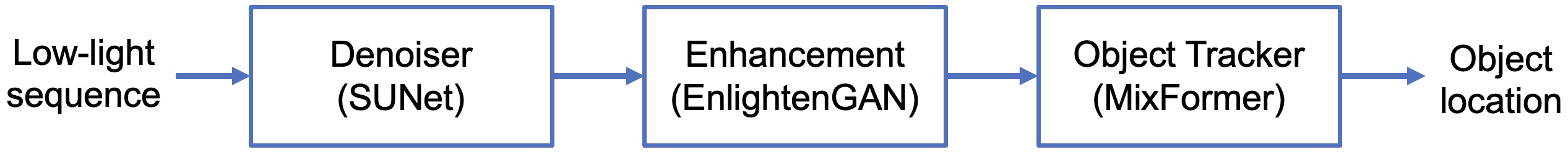}
\centering
\caption{The diagram used for our study on object tracking in low-light scene}
\label{fig:diagram}
\end{figure}

Despite the existence of specific methods proposed for low-light enhancement, we chose to adopt separate denoising and light correction approaches. This decision allows us to investigate the distinct impacts of various distortions on the tracker's performance within low-light environments. The workflow, depicted in Fig. \ref{fig:diagram}, integrates the MixFormer tracker with two preprocessing modules. Depending on the specific cases studied, one or both of these preprocessors might be omitted. Moreover, with the utilization of synthetic low-light datasets, we have access to clean, daylight ground truth data, enabling us to fine-tune the networks.

\subsection{Preprocessing with denoising}
\label{sec:denoise}
\noindent
In visual object tracking tasks, noise is inevitable and can significantly impact tracking efficiency. A common solution to address this issue is to preprocess tracking data before inputting it into the tracking network. Denoising techniques, such as filtering, temporal accumulation, and learning-based methods, are widely used in practice \cite{zhang2017denoising, guo2018real, Malyugina:topological:2023}.

In this paper, we adopt the state-of-the-art method, SUNet \cite{fan2022sunet} for denoising. This model, although simple, effectively combines the Swin Transformer and UNet architectures, enhancing feature extraction and hierarchical representation capabilities. Its dual up-sample block architecture, employing subpixel and bilinear up-sampling methods, helps prevent checkerboard artifacts and enhances overall performance. Demonstrating competitive results on widely-used denoising datasets, the SUNet model proves its practical effectiveness in addressing real-world image denoising issues, setting it apart from existing models. In this project, a pretrained SUNet model is utilized to preprocess the input dataset via denoising, aiming for improved tracking performance.

\subsection{Preprocessing with enhancement}
In the previous sections, a methodology was discussed for addressing noise in low-light sequences. However, other low-light features, such as color imbalance and low contrast, also contribute to the degradation of tracking performance. Various light enhancement methods have been proposed in the past, ranging from histogram based ones to learning based ones. 

Here, we adopt EnlightenGAN \cite{jiang2021enlightengan} which is a deep learning based generative adversarial network. The model represents a significant advancement in the field, introducing a pioneering unpaired training strategy that eliminates the need for paired training data and improves real-world generalisation. Its innovative global-local discriminator structure addresses spatially-varying light conditions effectively, while self-regularisation techniques, including self feature preserving loss and self-regularised attention mechanism, contribute to the model's success in the unpaired setting. EnlightenGAN offers superior performance and adaptability in comparison to state-of-the-art methods. From the diagram in Fig. \ref{fig:diagram}, when the EnlightenGAN is fine-tuned, the least-square GAN loss $L_G$  is applied to the generator of the EnlightenGAN.

\subsection{MixFormer}

MixFormer \cite{cui2022mixformer} tracks the target object by progressively extracting coupled features for the target template and search area while deeply integrating the information between them. This architecture consists of two main components: i) a backbone, which comprises iterative target-search MAMs (Mixed Attention Mechanism), and ii) a localization head, which is responsible for producing the target bounding box. The MAM blocks allow for the simultaneous extraction and integration of features from the target template and search area. The localization head simplifies the process of localising the tracked object within the search area, making the overall pipeline more efficient.

One of the key advantages of the MixFormer model is its compact and neat tracking pipeline. Unlike other trackers that typically decouple the steps of feature extraction and information integration, MixFormer combines these steps within its single backbone. This design choice results in a more efficient and streamlined architecture. Additionally, the MixFormer model does not require an explicit integration module or any post-processing steps, further simplifying the overall tracking pipeline. This simplification can lead to reduced computational complexity and faster inference times, making the MixFormer model a more suitable option for real-time tracking applications.

\vspace{2mm}
 \noindent \textbf{Mixed Attention Module} (MAM)  processes the input target template and search area with the aim of simultaneously extracting their long-range features and fusing the information interaction between them. This module enhances the tracker's ability to capture and integrate essential information from both the target and search area smoothly. 
Unlike the original Multi-Head Attention mechanism \cite{NIPS2017_3f5ee243}, the MAM operates on two separate token sequences corresponding to the target template and search area. It achieves this through dual attention operations. Self-attention is performed on the tokens (image patches) in each sequence (target and search) themselves to capture target or search specific information. Cross-attention is conducted between tokens from two sequences to allow communication between the target template and the search area. A concatenated token sequence is used to implement the mixed attention mechanism. Let vector $q_t$, $k_t$ and $v_t$ to represent target, $q_s$, $k_s$ and $v_s$ to represent search region. The mixed attention can be defined as:
\begin{equation}
    k_m = Concat(k_t,k_s) , \ v_m = Concat(v_t,v_s),
\end{equation}
\begin{equation}
   Attention_t = softmax(\frac{q_t k_{m}^T}{\sqrt{d}})v_m, 
\end{equation}
\begin{equation}
   Attention_s = softmax(\frac{q_s k_{m}^T}{\sqrt{d}})v_m, 
\end{equation}
\noindent
where $d$ denotes the dimension of the key vectors, $Attention_t$ and $Attention_s$ are the attention maps of the target and search respectively.

To achieve additional modeling of local spatial context, a separable depth-wise convolutional projection layer is performed on each feature map (i.e., query, key and value). Then each feature map of the target and search is flattened and processed by a linear projection to produce queries, keys, and values of the attention operation. Finally, the target token and search token are concatenated and processed by a linear projection.



\vspace{2mm}
\noindent \textbf{Online Template Update} plays a crucial role in capturing temporal information, as well as addressing object deformation and appearance variations in visual tracking. However, it is widely acknowledged that poor-quality templates may result in inferior tracking performance. Consequently, the authors introduce a score prediction module (SPM) to select reliable online templates based on their predicted confidence scores.

The SPM comprises two attention blocks and a three-layer perceptron. Initially, a learnable score token serves as a query to attend to the search ROI (region of interest) tokens. This process enables the score token to encode the extracted target information. Subsequently, the score token attends to all positions of the initial target token, implicitly comparing the extracted target with the first target. Finally, the score is generated by the MLP (multi-layer perceptron) layer and a sigmoid activation function.

The online template is considered negative when its predicted score falls below 0.5. By filtering out low-confidence templates, the SPM helps improve the overall tracking performance. The introduction of the SPM ensures that the tracker utilises high-quality templates for tracking, which in turn enhances its ability to adapt to object deformation and appearance changes. This approach enables more accurate and robust tracking performance in various challenging scenarios.

\vspace{2mm}
\noindent \textbf{Loss Function} used by the MixFormer model is a combination of L1 loss and GIoU loss. It is denoted as follows:
\begin{equation}
    L_{loc} = \lambda_{L1} L1 (B_i,\hat{B}) +\lambda_{GIoU} L_{GIoU} (B_i,\hat{B})\\
\end{equation}
where $\lambda_{L1} = 5$ and $\lambda_{GIoU} = 2$ are the weights of the two
losses, $B_i$ is the ground-truth bounding box and $\hat{B}$ is the predicted bounding box. L1 Loss is commonly used because of its robustness and insensitivity to outliers. Object tracking often involves dealing with occlusions, sudden motion changes, and noisy measurements. L1 Loss is less sensitive to outliers because it considers the absolute differences, thus it is more robust and makes it an ideal choice for visual object tracking tasks. 

Generalised Intersection over Union (GIoU) loss $L_{GIoU}$ is designed to address the limitations of the commonly used Intersection over Union (IoU) metric as it does not provide meaningful gradients for non-overlapping bounding boxes \cite{rezatofighi2019generalized}. GIoU loss addresses this issue by extending the IoU metric to account for the non-overlapping bounding boxes as well. It is computed as follows:
\begin{equation}
    GIoU = IoU - \frac{|C \setminus (A \cup B)|}{|C|} = \frac{| A \cap B |}{| A \cup B |} - \frac{|C \setminus (A \cup B)|}{|C|}
\end{equation}
where $A$ and $B$ are the prediction and ground truth bounding boxes, $C$ represents the area of the smallest enclosing box containing both boxes.

For the online training stage, a standard cross-entropy loss is used to train the SPM. It is defined as follows:
\begin{equation}
L_{score} = y_i \log(p_i) + (1-y_i) \log(1-p_i) \\
\end{equation}
where $y_i$ is the ground-truth label and $p_i$ is the predicted confidence score.

\section{Experiments and discussion}\label{sec2}

\subsection{Synthetic Low-Light Dataset}
We used the GOT-10K dataset \cite{got10k}, which is a large-scale, high-diversity benchmark for visual object tracking, comprising a wide variety of sources, such as YouTube, Vimeo, and Dailymotion. GOT10K contains more than 10,000 videos and covers 560 distinct object classes. The predefined testing set consists of 420 videos, including 84 different object classes and 31 forms of motion. To prevent larger-scale classes from dominating the evaluation results, the maximum number of videos for each class was limited to 8, which accounts for only 1.9\% of the test set size. The validation set was created by randomly sampling 180 videos from the training subset, with a uniform probability distribution across different object classes.

GOT-10K dataset was captured in normal light and good conditions, whereas video sequences taken in poor lighting conditions often display attributes like low brightness and contrast, a limited grayscale spectrum, color distortion, and considerable noise. To synthesize low light, we followed the image degradation model proposed in \cite{7351548} and we included the color imbalance effect $\mathcal{C}$  in the model as shown in Equation \ref{eqn:internsity}: 
\begin{equation} 
\label{eqn:internsity}
g(x,y) = \mathcal{C}\left(\alpha \cdot{f(x,y)^ \gamma}  + \beta\right) + \epsilon_n, 
\end{equation}

\noindent where $g(x,y)$ is the output image, $f(x,y)$ is the input image, $\alpha$ is the contrast adjustment parameter, $\beta$ is the brightness adjustment parameter, $\gamma$ is the gamma factor, and $\epsilon_n$ represents Gaussian noise. 

An $\alpha$ value above 1 boosts image contrast, darkening dark areas and brightening bright areas. An $\alpha$ value below 1 reduces image contrast, lightening dark areas and darkening bright areas. An $\alpha$ value of 1 maintains the image's contrast unchanged.  A positive $\beta$ increases image brightness, a negative $\beta$ decreases it, and $\beta$ at 0 maintains the brightness. The $\gamma$ value describes the nonlinearity of the imaging system to different input brightness levels. Typically ranging from 0.1 to 5, a gamma of 1 signifies a linear relationship between input and output brightness. A gamma above 1 accentuates sensitivity to darker areas, while below 1 emphasizes brighter areas.

To create a color imbalance effect $\mathcal{C}$ in dark images, it can be done by selectively manipulating the saturation channel ($S$) without altering the hue ($H$) or value ($V$) channels. This is achieved by applying a scaling factor to the saturation channel, which can be represented by the equation $S' = S \cdot \alpha_S$, where $S'$ is the adjusted saturation, $S$ is the original saturation, and $\alpha_S$ is the scaling factor. By selectively adjusting the saturation of specific color channels, an imbalance in color distribution can be created that mimics the appearance of color imbalance often observed in real-world low-light conditions. Note that modifying the $V$ channel alone will not achieve color imbalance, as it only affects the overall brightness of the image without altering the color relationships. We refrained from adjusting $H$ as it tended to alter white balance, a task already effectively addressed by commercial software.

Finally, we added Gaussian noise with the mean $\mu$=0 and the standard deviation $\sigma$, determining the spread or the variability of the noise added to the image. A larger standard deviation implies that the image is more ``grainy" or ``fuzzy" due to the presence of more random noise values. 

\subsection{Training setting}
\label{sec:stage2}
We trained the models with various synthetic low-light data with diverse parameters, including Gaussian noise ($\sigma$), gamma ($\gamma$) adjustment and saturation adjustment scaling factor ($\alpha_S$). These trackers, along with the tracker attained by normal light, were tested on a single synthesised dark test set to evaluate the tracking results of training with different parameters and to assess the impact of different low-light features on the tracking accuracy of the tracker.  The range of the parameters were set as follows:
\begin{itemize}
    \item For Gaussian noise, the mean value was constant at 128 for all datasets, while the $\sigma$ was set to 10, 25, 40, 55 and 70, with the default being 10.
    \item For contrast adjustment, the linear intensity factor  maintained at 0.4 for all datasets, while the $\gamma$ was set to 0.2, 0.3, 0.4, 0.5 and 0.6, with the default being 0.5.
    \item For saturation adjustment, the scaling factor $\alpha_S$ was set to 0.2, 0.3, 0.4, 0.5 and 0.6, with the default being 0.4.
\end{itemize}
It is worth noting that to assess the impact of each parameter on the tracking results, only one parameter was altered at each time, with the others set to their default values. Other training parameters remained the same as they were set for normal light to avoid the results being affected by other factors.

\subsection{Metrics}

\noindent \textbf{Intersection over Union (IoU)} is calculated as the ratio of the intersection of the predicted and ground-truth regions to their union. In other words, it measures the overlap between the two regions, where a value of 1 indicates a perfect match and a value of 0 indicates no overlap. 

\vspace{2mm}
\noindent \textbf{Area Under the Curve(AUC)}
refers to the area under the curve, plotting the fraction of successfully predicted frames against a threshold of IoU values. A higher AUC value suggests a better tracking performance as it indicates that the tracker is able to successfully track objects with larger IoU threshold $t$, which ranges from 0 to 1. The AUC  can be calculated using numerical integration and defined as follows:
\begin{equation}
    AUC = \int_{0}^{1} \frac{Number\:of\:frames\:with\:IoU >= t}{Total\:number\:of\:frames}  \,dt \\
\end{equation}

\noindent \textbf{OP50 and OP75} are the Overlap Percentages when the thresholds (as percentage) 50 and 75 are considered successful, respectively.  Typically, OP75 is considered a more strict criterion for measuring the performance as it sets a higher threshold for the overlap percentage. A higher OP50 or OP75 implies a better performance. 

\vspace{2mm}
\noindent \textbf{Precision} measures the accuracy of the predicted position of the tracked object. It calculates the average distance between the center of the ground truth bounding box and the center of the predicted bounding box for each sequence \cite{szeliski2022computer}. The Precision is the proportion of frames of which the distance is below a threshold $d$:
\begin{equation}
    precision(d) = \frac{Number\:of\:frames\:with\:distance_i <= d}{Total\:number\:of\:frames}\\
\end{equation}
\noindent where $distance_i$ is the Euclidean distance between the center points in frame $i$:
\begin{equation}
    distance_i = \sqrt{(x_{gt_i} - x_{pred_i})^2 + (y_{gt_i} - y_{pred_i})^2}
\end{equation}
where $x_{gt_i}$ and $y_{gt_i}$ are the coordinates of the center of ground truth bounding box, and $x_{pred_i}$ and $y_{pred_i}$ are the coordinates of the center of the predicted bounding box. 

\vspace{2mm}
\noindent \textbf{Normalized Precision} takes into account the differences in object sizes and frame resolution by normalizing the distance between the ground truth and predicted bounding box centers. The normalisation is commonly done by dividing the $distance_i$ by the diagonal length of the ground truth bounding box, which can be defined respectively as follows.\cite{szeliski2022computer}, where where $d$ is the threshold:
\begin{equation}
    Normalized\:Distance_i = \frac{distance_i}{Diagonal\:length\:of\:ground\:truth\:bounding\:box_i} \\
\end{equation}
\begin{equation}
    Normalized\:Precision(d) = \frac{Number\:of\:frames\:with\:Normalized\:distance_i <= d}{Total\:number\:of\:frames}
\end{equation}

\subsection{Impact of low-light distortions on tracking performance}

This section investigates the impact of individual distortions observed in low-light environments—such as noise, gamma, and saturation changes—on tracking performance. This demonstrates the parameters that should be set to generate synthetic low-light videos for training the model to achieve optimal performance when used in a general scenario.

\subsubsection{Noise levels}

\begin{figure*}[t]
     \centering     
     \includegraphics[width=.49\textwidth]{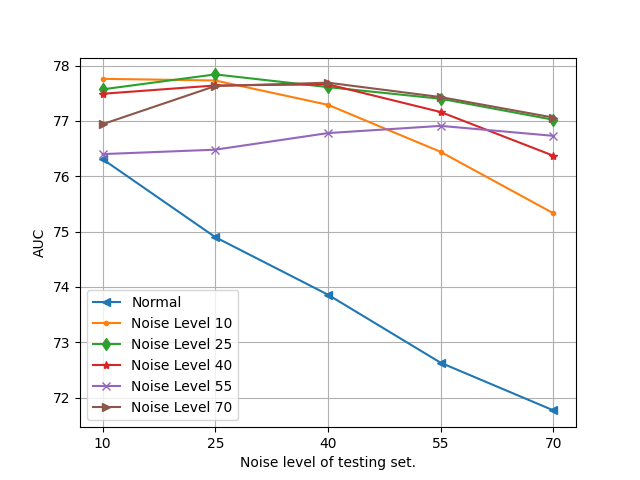}
     \includegraphics[width=.49\textwidth]{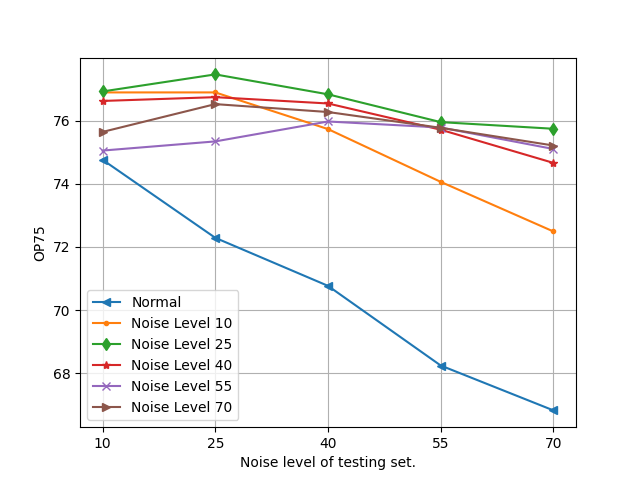}
     \includegraphics[width=.49\textwidth]{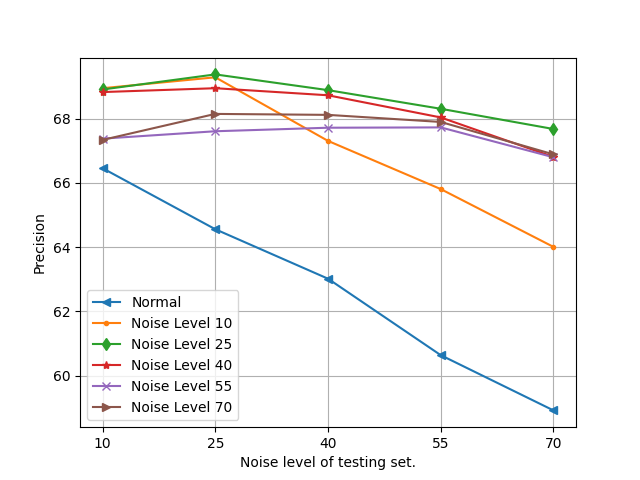}
     \includegraphics[width=.49\textwidth]{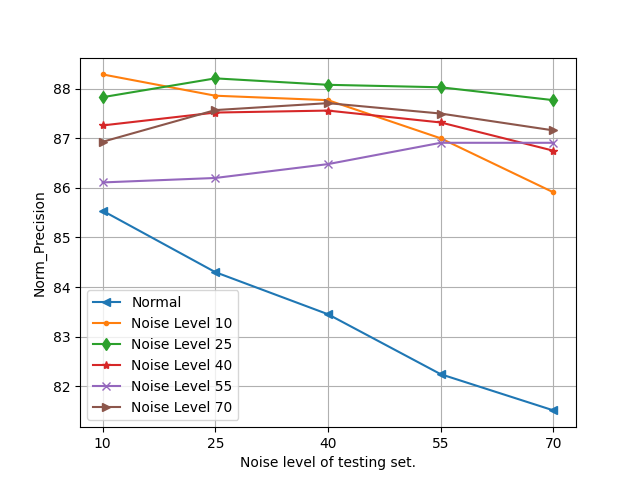}
     \caption{Test results of trackers trained with different \textbf{noise level}. The x axis shows the noise level of the test sets, while the Y axis shows the values of testing metrics.}
     \label{fig:sigma_test}
\end{figure*}

We explored the impact of varying noise levels on the tracker's performance. While maintaining normal lighting conditions, we adjusted noise levels by generating test sets with different sigma values: 10, 25, 40, 55, and 70 for each set. All other parameters were maintained at their default as specified in section \ref{sec:stage2}.  The results are shown in Figure \ref{fig:sigma_test}. When the model was trained in normal light without noise, it showed poor robustness to noise (as indicated by the blue line in the plots). Surprisingly, the model trained with a noise level of 25 demonstrates the highest performance in object tracking across varying noise levels. Conversely, models trained with higher noise levels failed to achieve optimal tracking performance, even when tested under similar noise conditions. This struggle may indicate that the network faces difficulties in capturing features from very noisy inputs.

\subsubsection{Gamma values}

\begin{figure*}[t]
     \centering     
     \includegraphics[width=.49\textwidth]{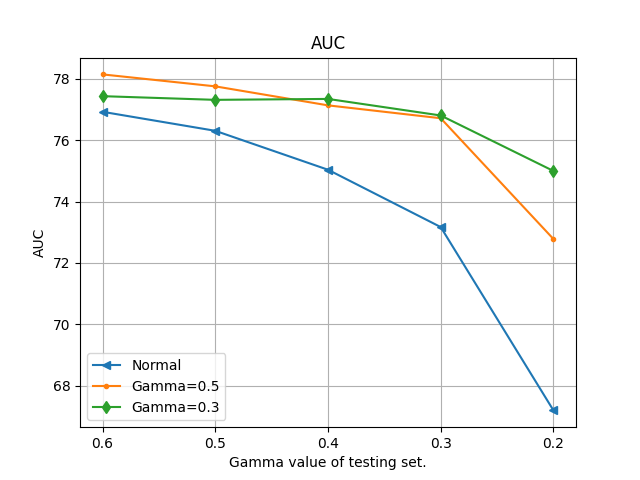}
     \includegraphics[width=.49\textwidth]{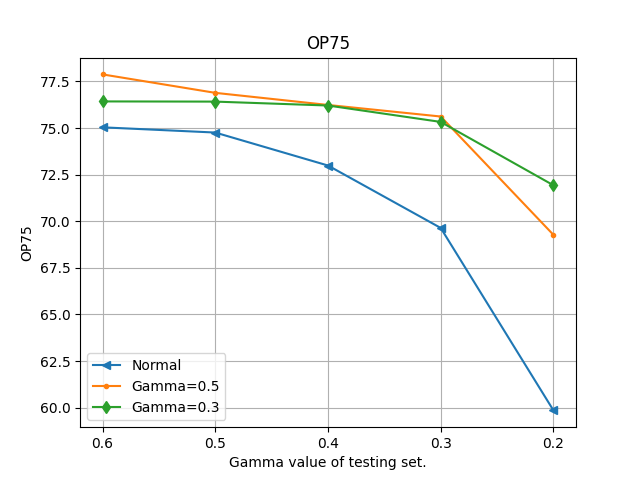}
     \includegraphics[width=.49\textwidth]{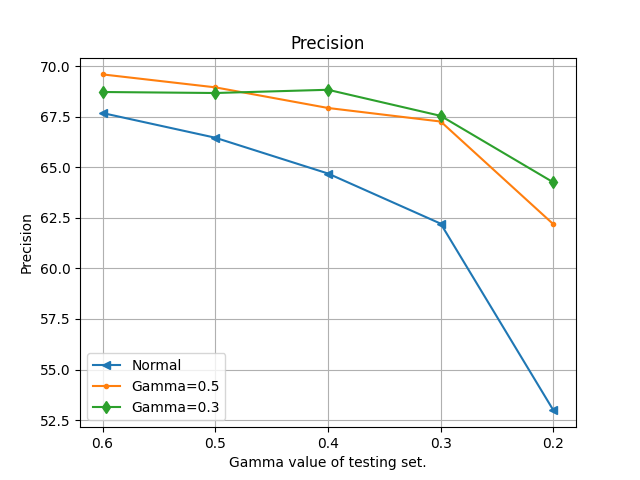}
     \includegraphics[width=.49\textwidth]{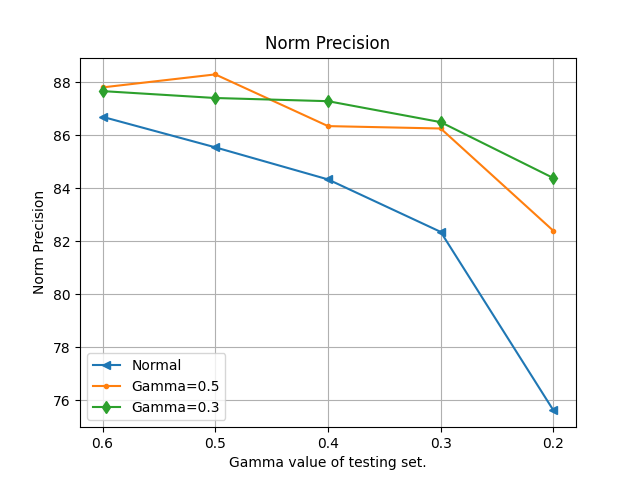}
     \caption{Test results of trackers trained with different \textbf{gamma gains}. The x axis shows the \textbf{gamma value} of the test sets, while the Y axis shows the values of testing metrics.}
     \label{fig:gamma_test}
\end{figure*}

\begin{figure*}[t!]
    \centering
    \includegraphics[width=.495\textwidth]{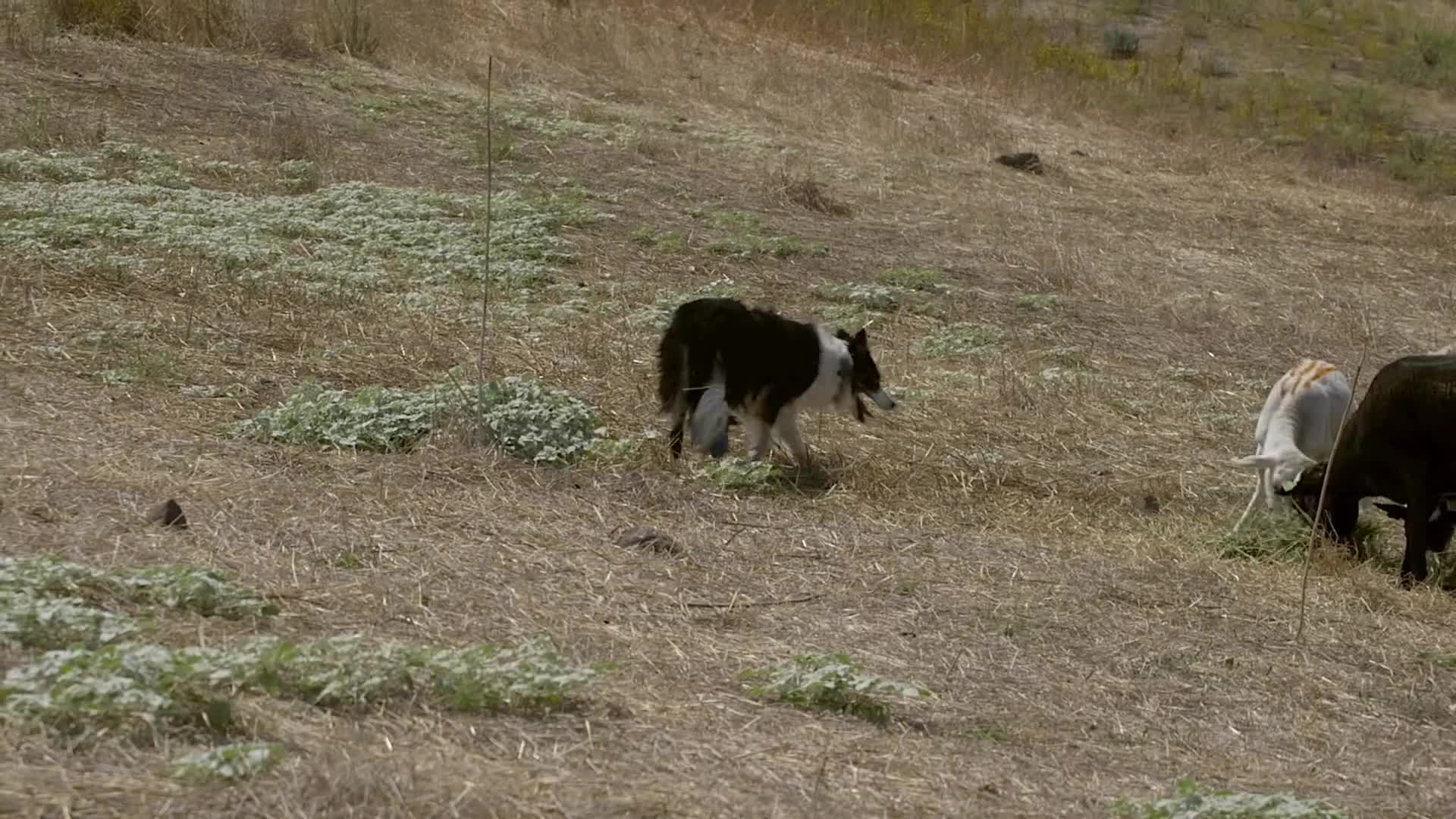}
    \includegraphics[width=.495\textwidth]{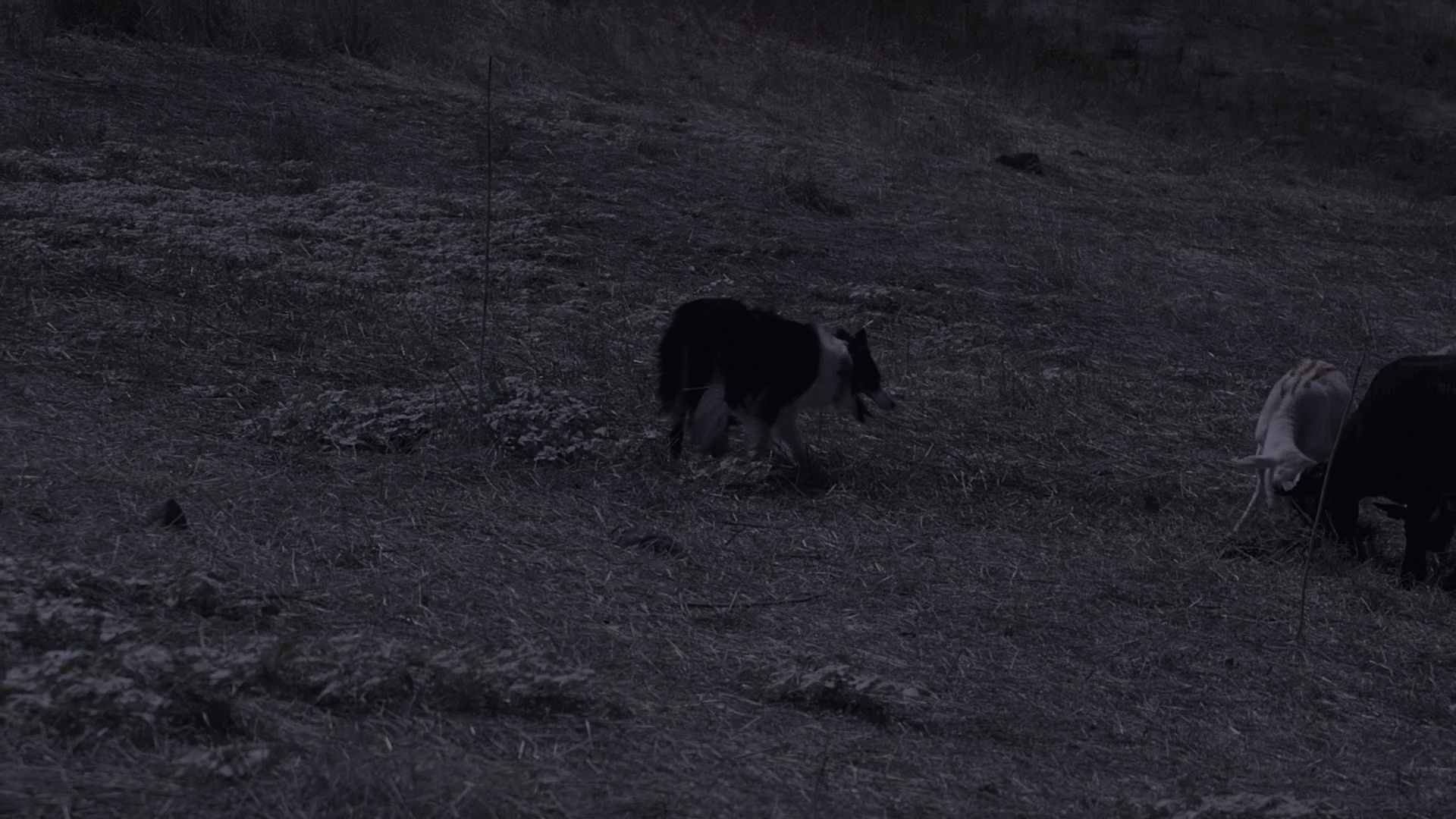} \\
    \includegraphics[width=.495\textwidth]{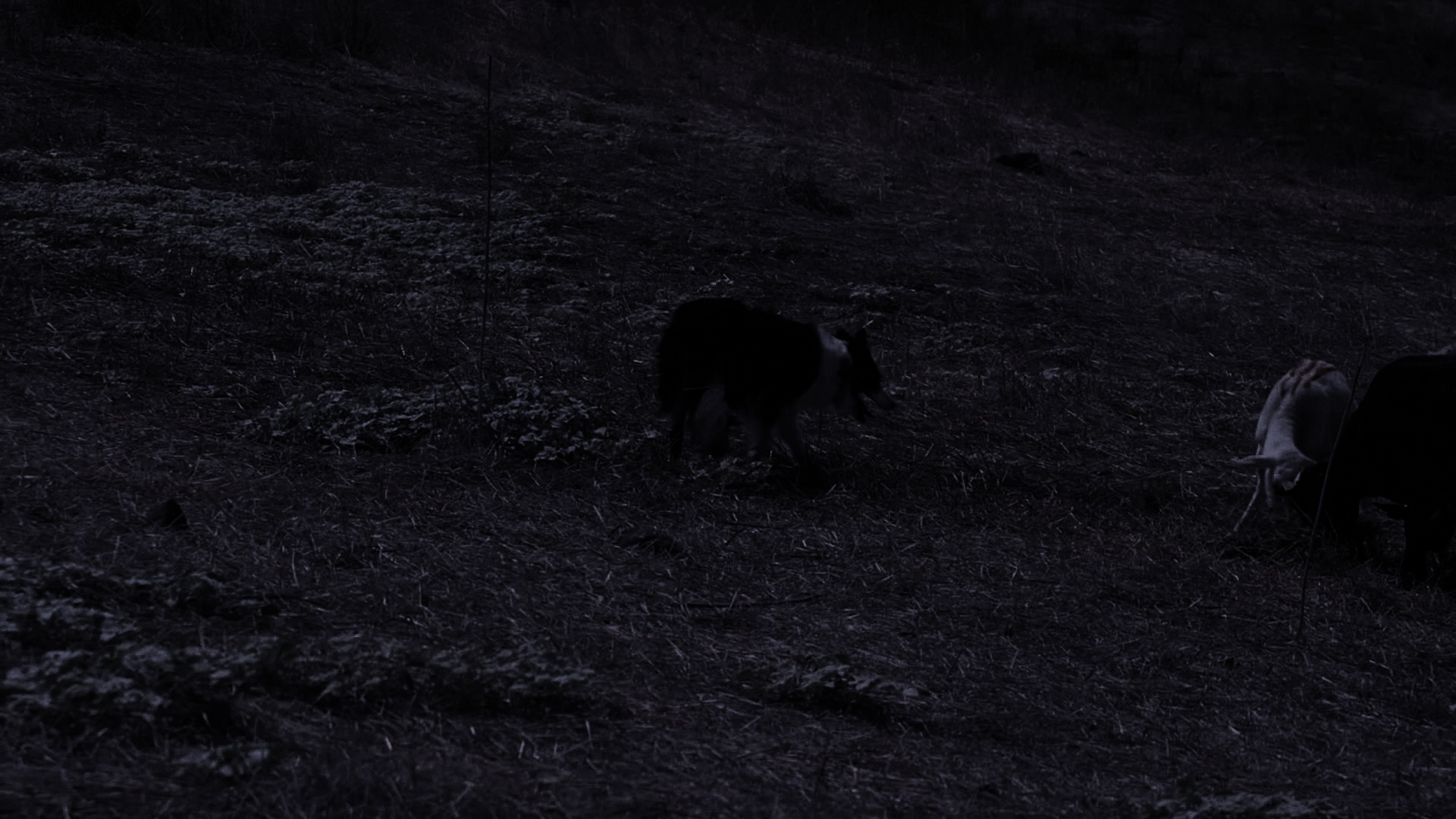}
    \includegraphics[width=.495\textwidth]{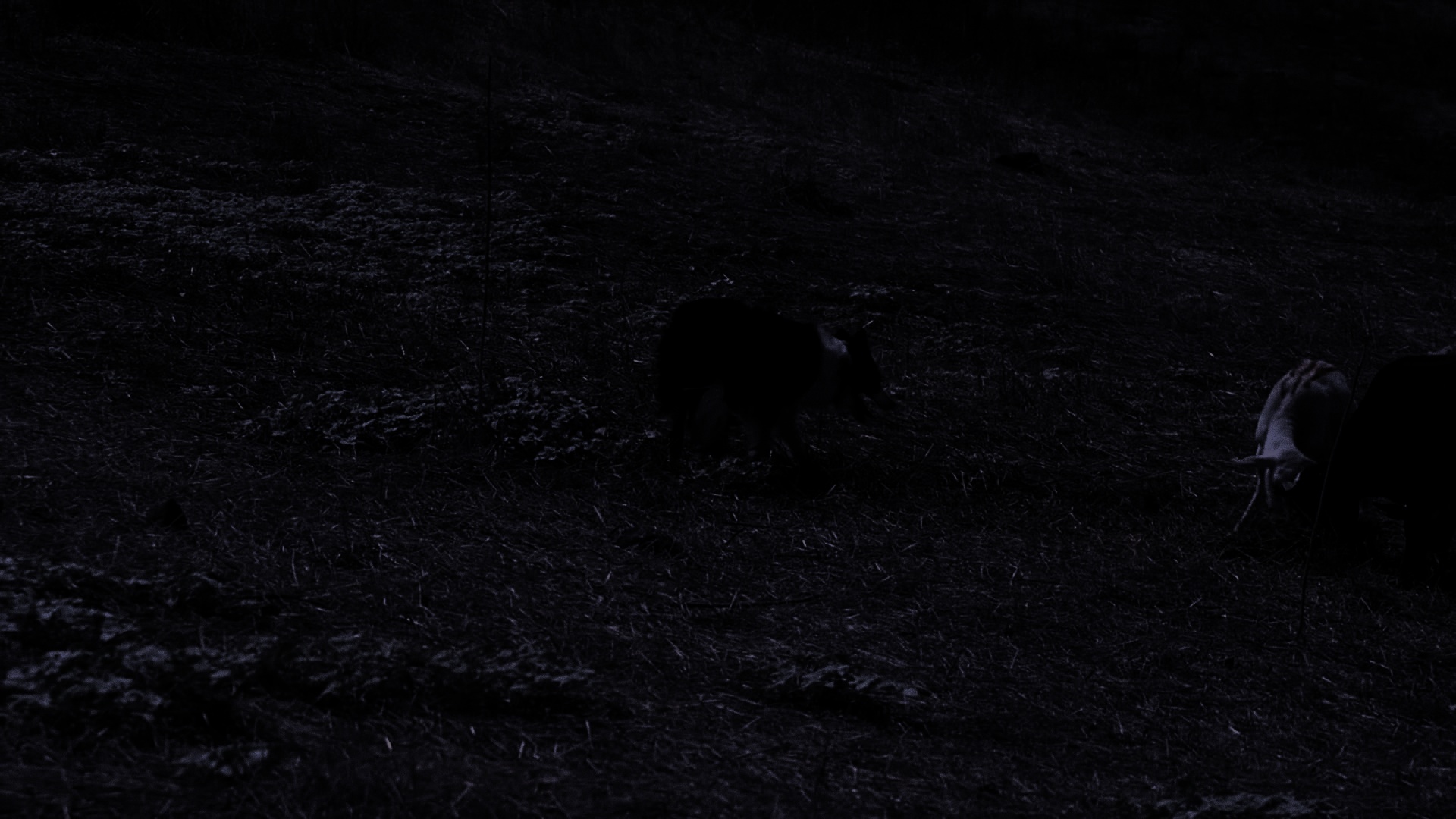}
    \caption{Images with different levels of \textbf{gamma gain}. Top left shows the image in original daylight environment. Other images has gamma gains of 0.6, 0.3 and 0.2 respectively.}
    \label{fig:gamma_example}
\end{figure*}

Figure \ref{fig:gamma_test} illustrates the test results of the three trackers trained with different gamma values. A notable observation is that the testing result of model trained with daylight dataset exhibits a non linear decrease. Specifically, in between gamma gain of 0.2 and 0.3 the model's precision experienced a sharp decline. The same trend can be found in trackers trained with gamma gain of 0.3 and 0.5. This phenomenon can be explained by the characteristics of gamma correction. When the gamma gain decreases to a certain level, the image becomes extremely dark, hence the object is not visually distinguishable and it is also difficult for machine to extract useful features. Unlike noise, which distorts edges and destroys certain features, low brightness and low contrast cause the object to blend into the background, making it impossible to identify edges or features. Figure \ref{fig:gamma_example} shows the original daylight image in comparison to the synthesised outcome of images with gamma gains 0.6, 0.3 and 0.2 respectively.

\begin{figure*}[t!]
     \centering     
     \includegraphics[width=.49\textwidth]{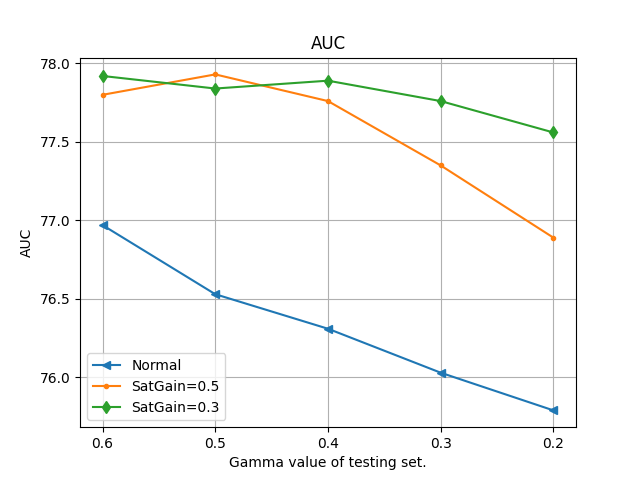}
     \includegraphics[width=.49\textwidth]{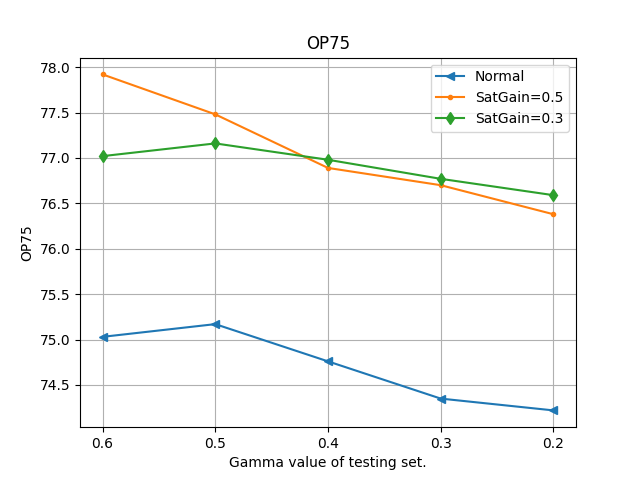}
     \includegraphics[width=.49\textwidth]{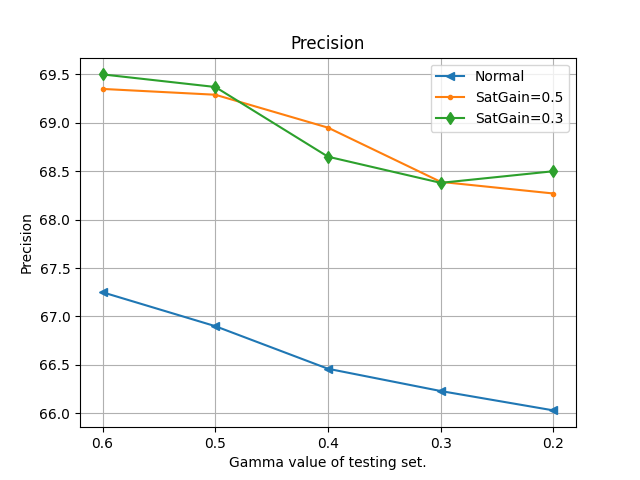}
     \includegraphics[width=.49\textwidth]{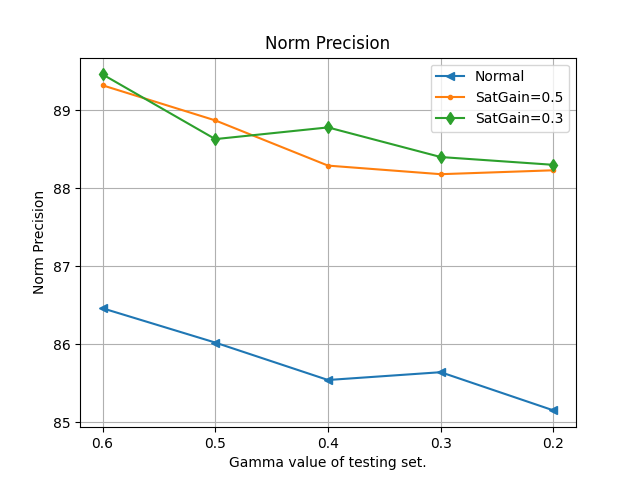}
     \caption{Test results of trackers trained with different saturation gains. The x axis shows the \textbf{saturation gain} values of the test sets, while the Y axis shows the values of testing metrics.}
     \label{fig:sat_test}
\end{figure*} 

\subsubsection{Saturation values}

Showing in Figure \ref{fig:sat_test}, a descending pattern can be found in the curves as the saturation gain reduces. Furthermore, the trackers trained with saturation gains of 0.3 and 0.5 display a significantly improved performance compared to the track trained on the daylight dataset. This observation aligns with the previously mentioned findings regarding the impact of noise and gamma gain, where trackers trained on the synthetic low-light dataset show better robustness when tested on various dark dataset. This consistency suggests that training the model on dataset with diverse low-light features can improve their versatility and effectiveness when handling visual object tracking tasks in a wide range of low-light scenarios. 

The impact of saturation on the model's tracking efficiency is relatively smaller than that of noise and gamma gain. As the saturation gain drops from 0 to 0.2, these five metrics only decrease by approximately 3\%. For the impact of noise, the AUC decreases from 79.30\% to 73.86\% as the noise level rises from 0 to 40. Similarly, OP50, OP75, Precision, and Normalized Precision decline by 5.69\%, 9.55\%, 9.37\%, and 5.48\% respectively. Similarly, when the gamma value drops from 1 to 0.3, while the features remain recognizable, the AUC, OP50, OP75, Precision, and Normalized Precision decrease by 6.13\%, 6.8\%, 10.68\%, 10.08\%, and 6.58\% respectively. 

 While changes in saturation can alter the appearance of an image by adjusting the color intensity, they do not have massive impact on the overall image quality, or the visibility of the objects and their edges. This implies that alterations in shapes and edges deteriorate tracking performance more significantly than saturation changes do. Thus, the model is able to maintain high performance and is less affected by the changes in saturation compared to noise and gamma, where the features in the object are greatly impacted.

\subsection{Performance improvement with denoising and image enhancement}

Table \ref{tab:compare_600} shows the test results of trackers trained using synthetic dark datasets (sigma = 40, gamma = 0.5, saturation = 0.4), denoised datasets, and enhanced datasets. These trackers were then tested on dark, denoised, and enhanced data, respectively. The outcomes on the denoised dataset surpass those on the enhanced dataset, confirming that pre-processing with denoising significantly enhances the model's performance compared to enlightening. Moreover, upon applying denoising techniques to both the training and testing sets, the AUC increased by 5.36\% compared to training and testing on the original dark dataset. Conversely, applying preprocessing solely to the testing set resulted in a 3.21\% increase in AUC. This suggests that applying preprocessing to both training and testing procedures yields a more substantial improvement in the model's performance.

\begin{table}[t!]
    \caption{Test results of trackers trained on \textbf{dark, denoised and enhanced datasets} and tested on corresponding test sets. }
    \centering
    \begin{tabular}{ @{}lccccc@{}}
         \toprule
         Trackers & AUC & OP50 & OP75 & Precision & Norm Precision\\
         \midrule
         Dark & 61.29 & 73.43 & 57.60 & 51.23 & 72.84\\
         Denoised & 66.65 & 76.33 & 59.98 & 53.01 & 74.43\\
         Enhanced & 64.32 & 74.93 & 58.56 & 52.25 & 73.21\\
         Denoised+Enhanced & 67.15 & 77.12 & 60.72 & 53.68 & 75.18\\
         \botrule
    \end{tabular}
    \label{tab:compare_600}
\end{table}

\subsection{Visualised Tracking Results}

In this section, the visualised tracking results are discussed to further investigate the model's ability in handing challenging conditions. Specifically, the reasons to why the tracking failed in certain cases are examined to provide insights into the future improvements of the model. The main reasons for the model's failure to track objects are classified into three categories: i) ambiguity caused by the background, ii) presence of multiple, visually similar objects within the scene, and iii) occlusion or obstruction of the object. 

\subsubsection{Ambiguity caused by the background}
The background in the image can sometimes have similar features as the object. In low-light images, the visibility of the edges and textures of the object is degraded, making it more challenging for the tracker to distinguish between the object and the background when they share similar features, hence leading to tracking errors. 

\begin{figure*}[t!]
    \centering
    \includegraphics[width=.49\textwidth]{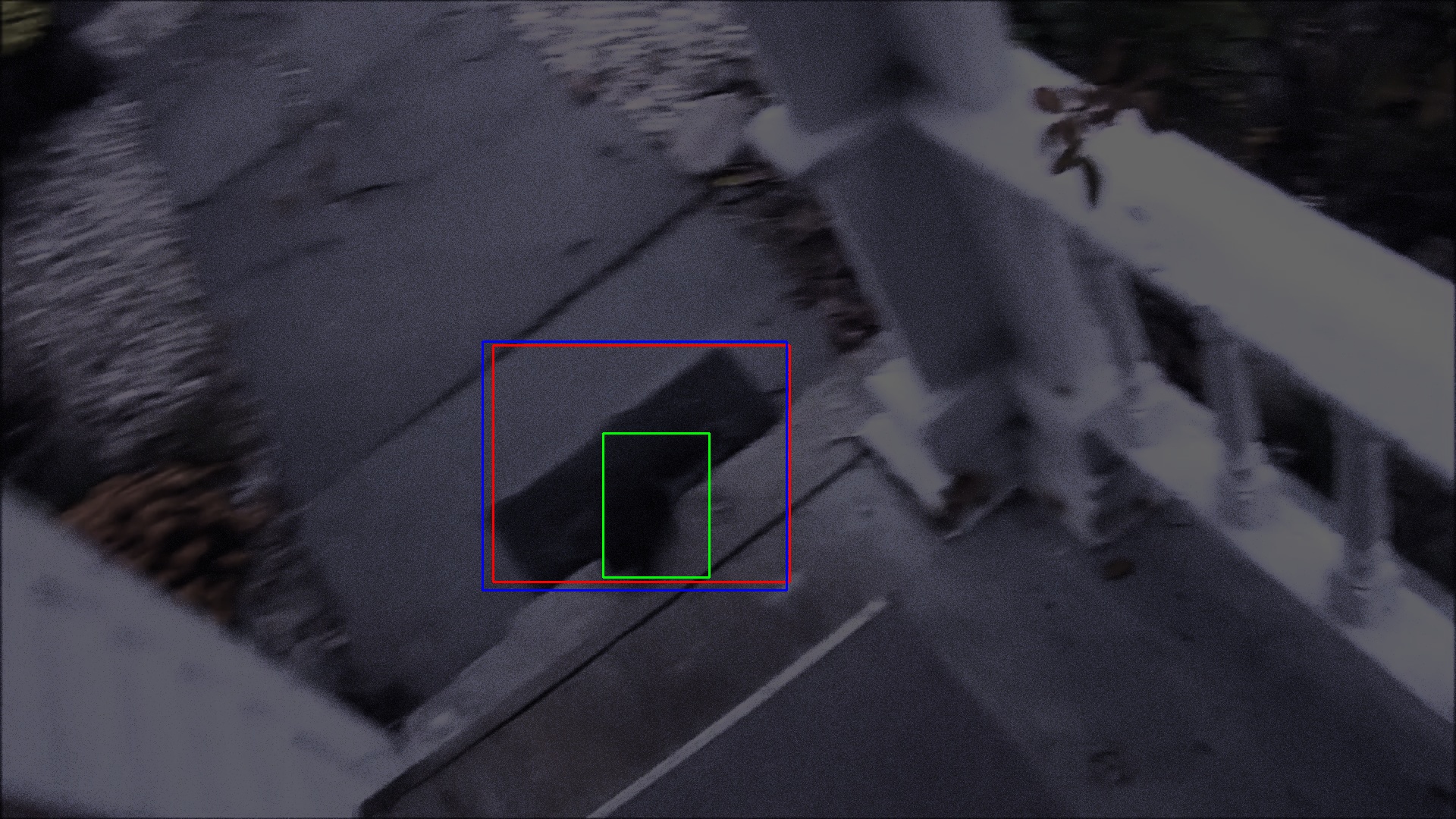}
    \includegraphics[width=.49\textwidth]{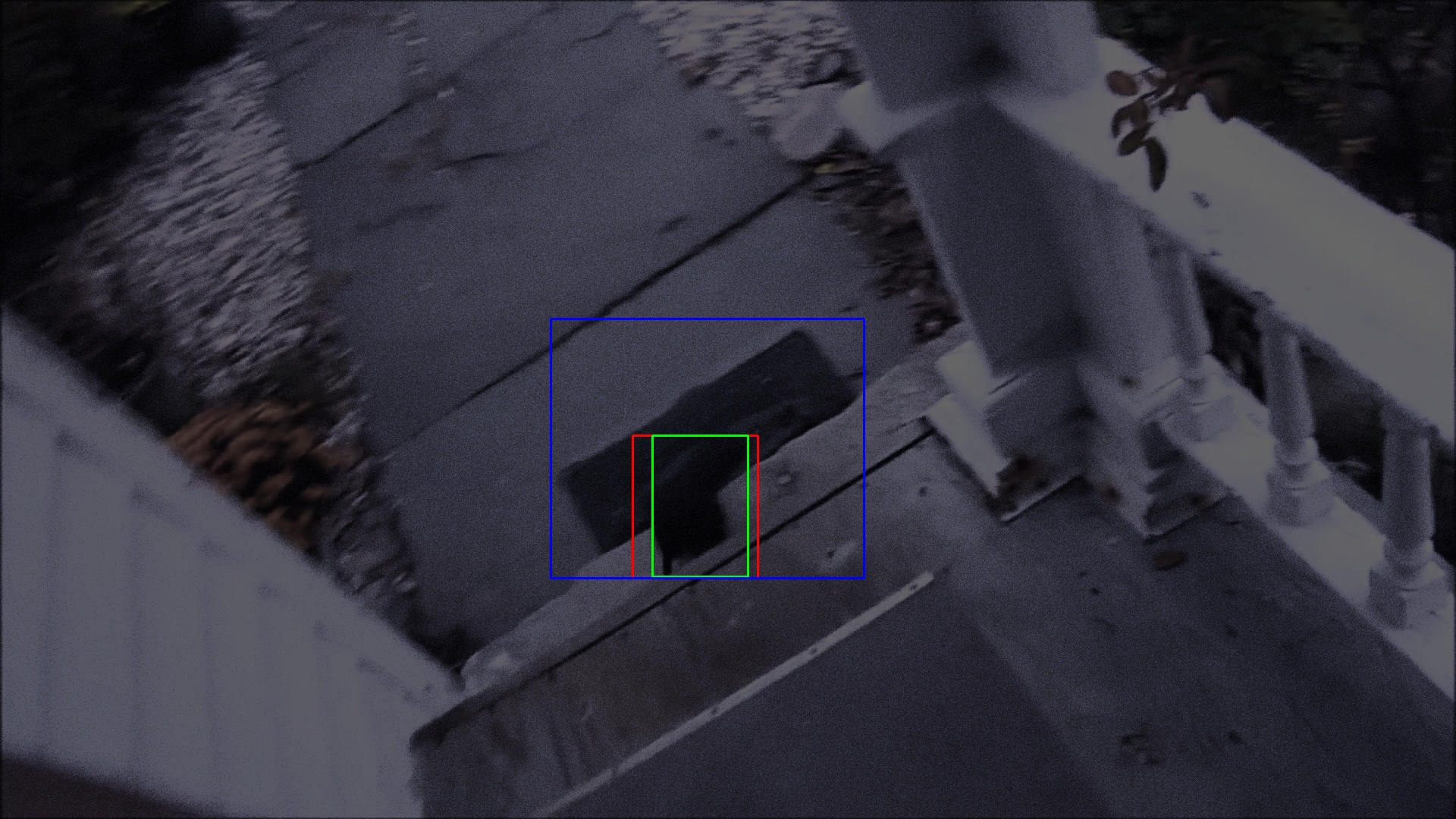}
    \caption{The image on the left show the example when both trackers fail to track a black cat in the dark due to confusion caused by background. The image on the right shows improved. Green, blue and red boxes are ground truth and the results from the models trained by daylight and dark datasets, respectively. }
    \label{fig:fail_background}
\end{figure*}

Figure \ref{fig:fail_background} displays two tracking failures, where the tracker struggles to differentiate the black squirrel from the background. The left shows when both the normal and dark trackers fail to identify the object in the frame. The cause of this error is that the door mat is mistaken as the squirrel, as they both appear black and have a slender shape. However, as presented in the image on the right, after the squirrel moves to the doorstep, where its features contract with the background, more features are captured by the dark tracker, hence it is able to correct the tracking result. On the contrary, the normal tracker continues to fail on recognising the object in this frame, indicating that its ability in feature extraction in low-light conditions is weaker than that of the dark tracker.

\subsubsection{Multiple Objects}

The challenges include consistently recognizing individual objects when multiple objects are present in a scene, managing interactions between objects, and coping with the appearance changes of each object.

\begin{figure*}[t]
    \centering
    \includegraphics[width=.49\textwidth]{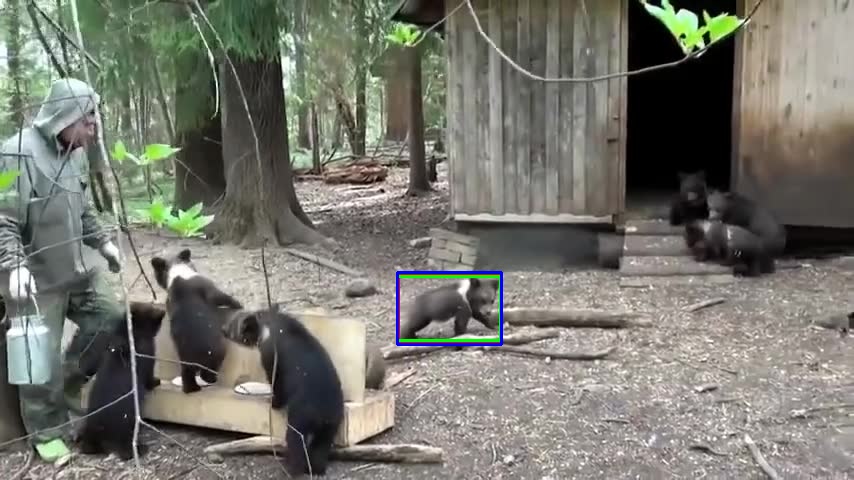}
    \includegraphics[width=.49\textwidth]{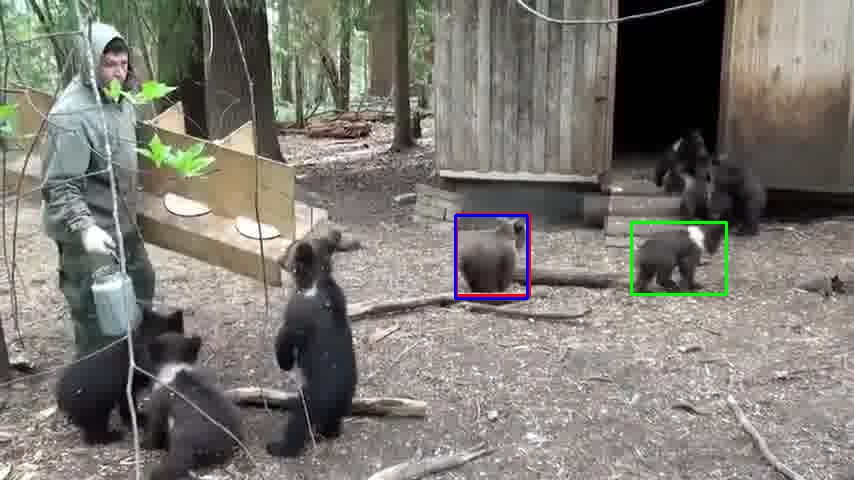}
    \caption{Examples when the normal tracker or both trackers failed to track an object in the dark due to multiple objects existing in the scene. Green, blue and red boxes are ground truth and the results from the models trained by daylight and dark datasets, respectively. The images are shown in the normal light for better visualisation.}
    \label{fig:fail_multi}
    \centering
    \includegraphics[width=.49\textwidth]{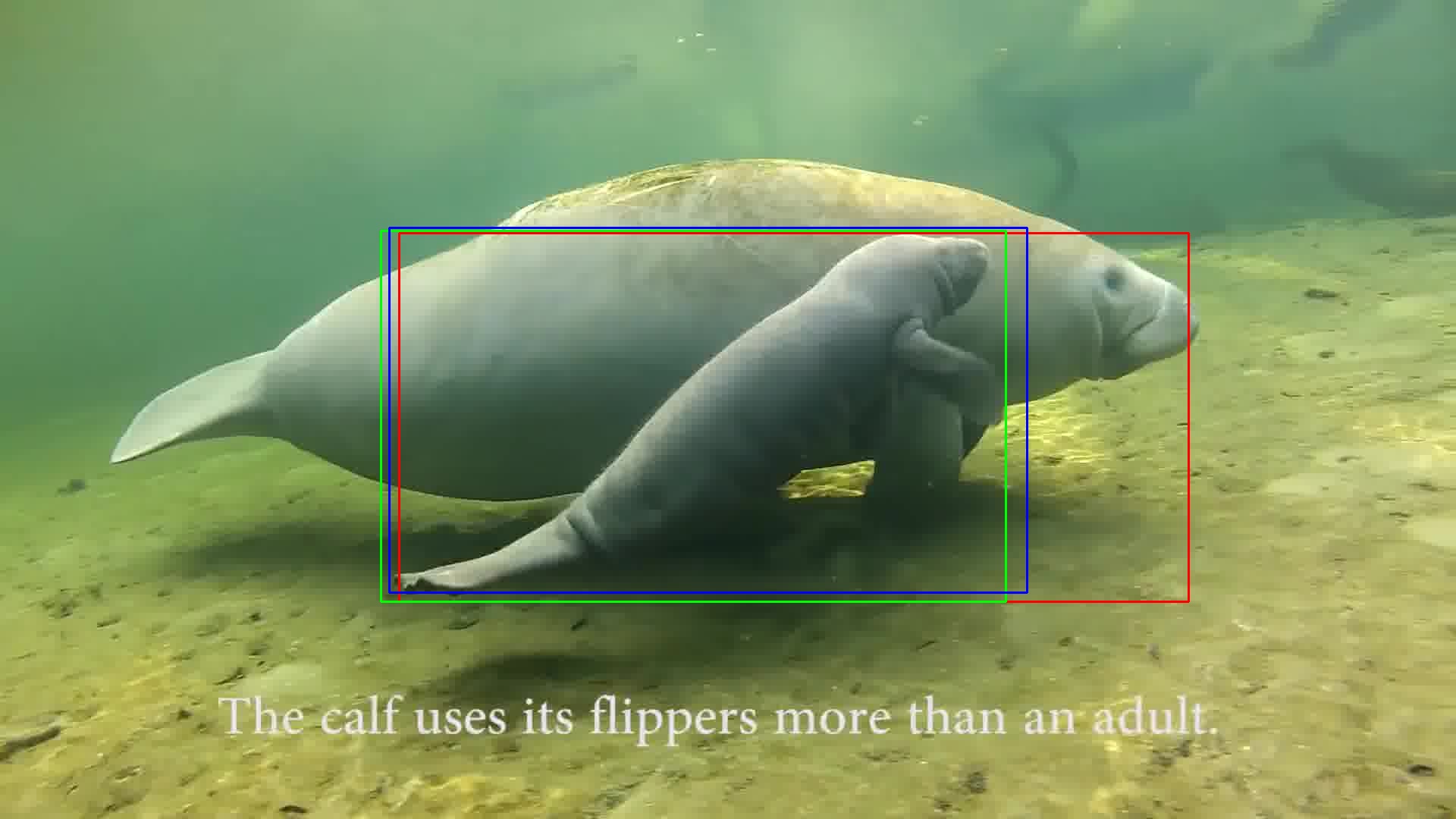}
    \includegraphics[width=.49\textwidth]{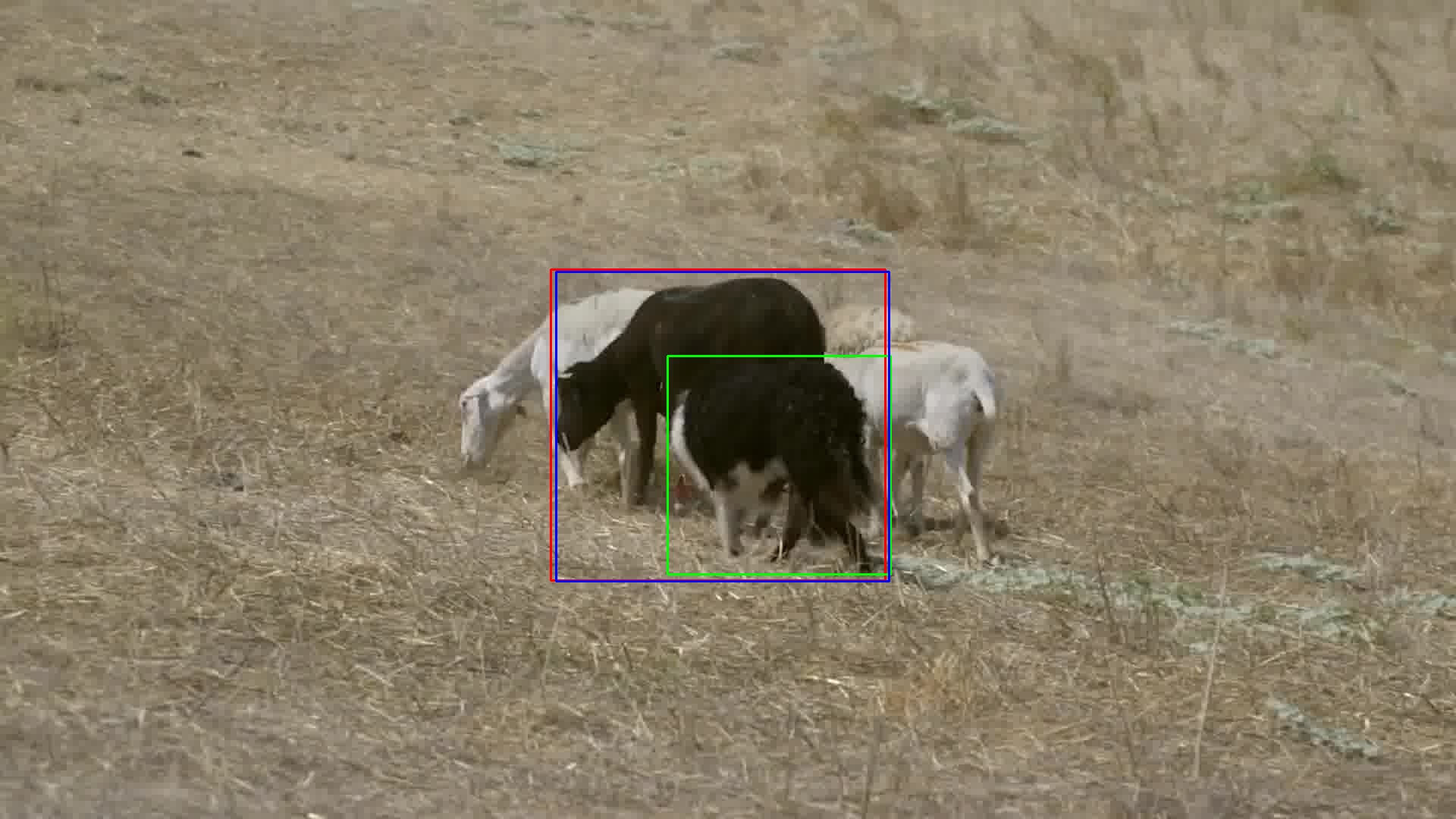}
    \caption{Examples when one or both the trackers fail to track an object in a frame due to interaction of objects. Green, blue and red boxes are ground truth and the results from the models trained by daylight and dark datasets, respectively. The images are shown in the normal light for better visualisation.}
    \label{fig:fail_multi_interact}
\end{figure*}

Figure \ref{fig:fail_multi} presents an example where the trackers are unsuccessful in maintaining the identity of the object. In the two frames, both trackers trained by normal light and low light manage to identify the object (as shown on the left) – a small black bear – when there is only one such bear walking on the ground. Nevertheless, in the right image, both trackers incorrectly identify the original black bear (bear 1) on the right and instead misinterpret the one on the left (bear 2) as the initial bear. This may occur due to the trackers' inability to accurately follow the object's movements. Specifically, as bear 1 moves to the right, bear 2 takes its original position. Consequently, despite the trackers' initial success in tracking the object and the minor changes in the visual appearance of bear 1, they still confuse bear 2 with the originally tracked bear 1. This highlights the trackers' lack of ability in accurately capturing the temporal information within the sequence. This limitation seems reasonable, considering the score prediction head in the original model design, which is crucial for capturing temporal information, is trained in the online stage of MixFormer. However, to reduce training complexity in this project, the online training step is excluded when training these trackers. As a result, the trackers may exhibit a diminished ability to capture temporal information.

Figure \ref{fig:fail_multi_interact} displays further examples where the trackers fail to track an object due to the presence of multiple objects in the scene. In both cases presented in the frames, the target object is interacting with another object in the scene. Consequently, some features of the other object involved in the interaction are incorrectly attributed to the target object. For example, in the left image, as the calf manatee interacts with the adult manatee,  the dark tracker falsely includes the adult manatee's head as part of the calf. This is probably because the head of adult manatee has more distinct features, such as the eyes and head shape, which the dark tracker can easily capture. Similarly, in the right image, the border collie is interacting with the black sheep. Since the border collie is in a position where its head is not visible in the picture, the sheep's head and neck, which have more distinguishable and pronounced features, are mistakenly identified by the trackers as part of the border collie. This mistake can also be attributed to the missing edges of the border collie's head, making it difficult for the tracker to identify the boundary of the object.

\subsubsection{Occlusion}

\begin{figure*}[t]
    \centering
    \includegraphics[width=.49\textwidth]{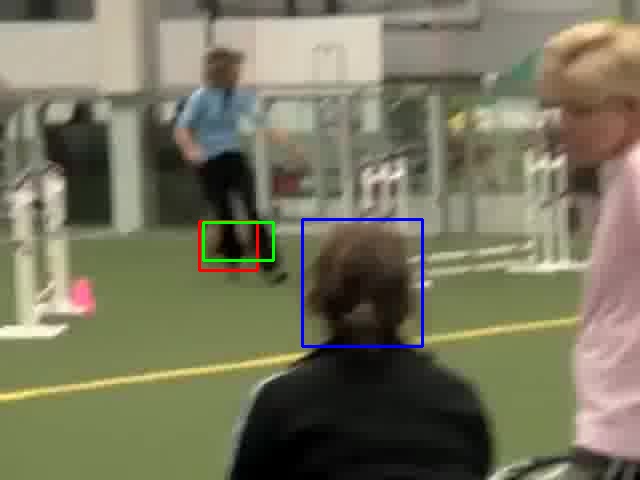}
    \includegraphics[width=.49\textwidth]{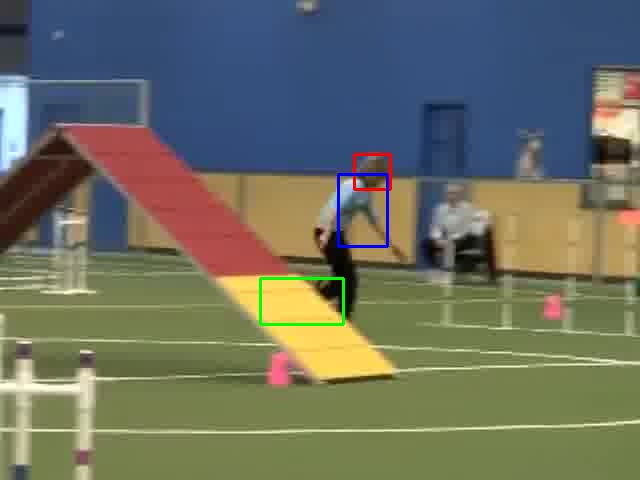}
    \caption{Left: When the object, a black dog, is partially occluded, the dark tracker is able to identify the part which is visible, while the normal tracker cannot identify the object. Right: when the object is fully obstructed, both trackers are unable to track the object in the frame. Green, blue and red boxes are ground truth and the results from the models trained by daylight and dark datasets, respectively.}
    \label{fig:fail_occlusion}
\end{figure*}

Occlusion poses a considerable challenge in object tracking, as it can impede the tracker's capacity to maintain a precise representation of the object throughout a sequence \cite{yilmaz2006object}. An illustration of this issue can be found in Figure \ref{fig:fail_occlusion}, where the trackers struggle to track an object, a black dog, during occlusion events. On the left, the object, a dog, is partially obscured by a person's legs. In this frame, the dark tracker is able to capture the object's features and define its edges despite the occlusion. In contrast, the daylight tracker inappropriately identifies the person's head, which shares similar features, such as brown and fuzzy appearance, with the dog in the image, as the target. This observation aligns with the previous finding that the dark tracker is more capable at capturing features and defining edges in low-light conditions than the normal tracker, allowing it to identify the object even when occlusion occurs in the dark. 

The dark tracker's ability to handle occlusion has its limitations. In the right frame, where the dog is entirely obscured by the structure, both trackers fail to recognise the animal. This failure can potentially be attributed to the model's inability to handle temporal information effectively due to the absence of online training, as previously discussed. When a model is adept at processing temporal information, it can leverage the object's motion patterns, trajectory, and appearance changes observed in previous frames to make predictions about the object's position and appearance during occlusion \cite{kalal2010forward}. Hence, when the model lacks this ability, it may fail to continuously track the object when the object is partially or entirely hidden from the view.

\subsection{MixFormer versus Siam R-CNN}
\label{sec:MixVSSiam}
This section compares our MixFormer-based tracker with Siam R-CNN \cite{siamrcnn}, a state-of-the-art object tracking model that merges Siamese networks with Region-based Convolutional Neural Networks (R-CNN). Siam R-CNN excels in robust and accurate visual tracking by effectively matching the target object across frames using a Siamese architecture. The R-CNN component aids in precise object localization and classification. This fusion enhances tracking performance, especially in challenging scenarios involving occlusions, deformations, and appearance changes. The following segment illustrates the performance comparison between these two methods.

Table \ref{apex:comp_test} displays the outcomes from testing sets generated using various parameters (e.g., sigma, gamma, saturation gain), while other settings remain default (see Section \ref{sec:stage2} for an in-depth explanation of parameter configurations). A noticeable observation in Table \ref{apex:comp_test} is the consistent lower performance of the Siam R-CNN model compared to the MixFormer model in both daylight and low-light tracking scenarios. This observation emphasizes the effectiveness of the MixFormer's MAM architecture, significantly enhancing tracking performance even under challenging lighting conditions. These results underscore the superiority of the transformer-based architecture over the conventional CNN network and underscore the advantages of the Mixed Attention Module.

\begin{table}[h]
\caption{Test results of MixFormer and Siam R-CNN on different test sets}\label{apex:comp_test}%
\begin{tabular}{@{}lcccc@{}}
\toprule
& \multicolumn{2}{@{}c@{}}{AUC} & \multicolumn{2}{@{}c@{}}{Norm Precision} \\\cmidrule{2-3}\cmidrule{4-5}%
Setting & MixFormer & Siam R-CNN & MixFormer & Siam R-CNN \\
\midrule
Normal & 79.30 & 77.21 & 88.93 & 86.83\\
         Sigma=10 & 76.31 & 74.92 & 85.54 & 84.02\\
         Sigma=40 & 73.36 & 70.24 & 83.45 & 81.73\\
         Gamma=0.3 & 73.17 & 71.58 & 82.35 & 80.57\\
         Saturation=0.3 & 76.18 & 74.86 & 85.79 & 85.13\\
\botrule
\end{tabular}
\end{table}

\section{Conclusion}
This paper examines the performance of object tracking algorithms in low-light conditions. The strategies involve training the model using synthetic datasets and applying denoising and image enhancement techniques in preprocessing. Our findings demonstrate that training the model on synthetic dark datasets notably improves its performance in low-light settings, particularly under varying noise and brightness levels. Our comprehensive study on the effects of low-light distortions reveals that noise has the most detrimental impact on tracking performance, followed by non-linear brightness changes. Training the model with a noise level of 25 and a gamma of 0.3 yields the best overall performance across various low-light conditions.
Additionally, we propose and evaluate two preprocessing methods, SUNet for denoising and EnlightenGAN for image enhancement, to enhance tracking accuracy. Implementing both techniques on the test set results in a 4.41\% improvement (AUC) in tracking accuracy compared to performance on the noisy dark dataset. Furthermore, utilizing denoising for both training and testing stages on the dark dataset leads to a 5.36\% improvement in tracking accuracy compared to models trained and tested on the original dark dataset.

\subsection*{Funding Information}
This work was supported by UKRI MyWorld Strength in Places Programme (SIPF00006/1), BRISTOL+BATH CREATIVE R+D (AH/S002936/1).

\subsection*{Data availability}
The datasets generated and analysed during the current study are available from the corresponding author on reasonable request.

\subsection*{Competing interests}
The authors declare that we have no known competing financial interests or personal relationships that could have appeared to influence the work reported in this paper.

\subsection*{Author contribution}
All authors contributed to the study conception and design. Material preparation, data collection and analysis were performed by Anqi Yi. The first draft of the manuscript was written by Anqi Yi and all authors commented on previous versions of the manuscript. All authors read and approved the final manuscript.

\subsection*{Research Involving Human and/or Animals}
Not Applicable

\subsection*{Informed Consent}
Not Applicable

\bibliography{sn-bibliography}

\end{document}